\documentclass[10pt,twocolumn,letterpaper]{article}

\usepackage[pagenumbers]{cvpr} 

\definecolor{cvprblue}{rgb}{0.21,0.49,0.74}
\usepackage[pagebackref,breaklinks,colorlinks,allcolors=cvprblue]{hyperref}

\def\paperID{13918} 
\def\confName{CVPR}
\def\confYear{2025}

\title{Post-Hoc MOTS: Exploring the Capabilities of Time-Symmetric Multi-Object Tracking}

\author{Gergely Szabó\\
{\tt\small szabo.gergely@itk.ppke.hu}
\and
Zsófia Molnár\\
{\tt\small molnar.zsofia@itk.ppke.hu}
\and
András Horváth\\
{\tt\small horvath.andras@itk.ppke.hu}
\and
ITK, PPCU\\
Budapest, Pr\'ater st. 50/A, 1083, Hungary
}

\begin{document}
\maketitle

\begin{abstract}

Temporal forward-tracking has been the dominant approach for multi-object segmentation and tracking (MOTS). However, a novel time-symmetric tracking methodology has recently been introduced for the detection, segmentation, and tracking of budding yeast cells in pre-recorded samples. Although this architecture has demonstrated a unique perspective on stable and consistent tracking, as well as missed instance re-interpolation, its evaluation has so far been largely confined to settings related to videomicroscopic environments. In this work, we aim to reveal the broader capabilities, advantages, and potential challenges of this architecture across various specifically designed scenarios, including a pedestrian tracking dataset. We also conduct an ablation study comparing the model against its restricted variants and the widely used Kalman filter. Furthermore, we present an attention analysis of the tracking architecture for both pretrained and non-pretrained models. All related codes, data, and models are available at \url{https://drive.google.com/drive/folders/1JbCJT4DMnzMIchqqC1IcKRjmYA0s5-IZ?usp=sharing}.

\end{abstract}
\vspace{-8pt}

\section{Introduction} \label{sec:intro}

Simultaneous segmentation, and tracking of multiple objects in image sequences remains a challenging task, requiring the optimization of many competing factors during the design of new methods. Most state-of-the-art trackers prioritize latency, as they are intended for use in real-time applications such as self-driving cars and surveillance cameras. However, these methods often overlook potential information gains and generalization options when inference time is not a critical parameter for the given application. A pertinent example is the tracking of living cells in videomicroscopic recordings, for which a novel architecture was designed and presented in the paper of Szabó \etal~\cite{szabo2023enhancing}, referred to hereafter as the \textit{"TS"} architecture due to its time-symmetric tracking approach.

While the original analysis demonstrates the outstanding capabilities of the architecture, training requirements, and hyperparameter dependencies specifically for instance segmentation and tracking of budding yeast cells and other synthetic cell-like objects in various scenarios, as well as its unique reconstructive capabilities that enable continuous tracking of objects --- a necessity in cell tracking due to the assignment challenges during cell division --- it does not fully explore the architecture's potential in substantially different environments.

Therefore, in this paper, we aim to extend the performance analysis using improved and more standardized metrics, conduct a comparative evaluation of the \textit{TS} architecture against the widely used Kalman filter \cite{KalmanFilter} and restricted variants of the \textit{TS} architecture, assess the architecture's performance in various specifically designed synthetic scenarios, and evaluate its performance in the challenging zero-shot knowledge transfer scenario between the synthetic MOTSynth-MOTS-CVPR22 training dataset \cite{MOTSynth} and the real-world person tracking MOTS dataset \cite{MOTS}. Furthermore we present an analysis of the local tracking segment of the \textit{TS} architecture using saliency maps to reveal its temporal spatial attention preferences.

\section{Architecture overview} \label{sec:Architecture}

\begin{figure*}[h!]
    \centering
    \includegraphics[width=.96\textwidth]{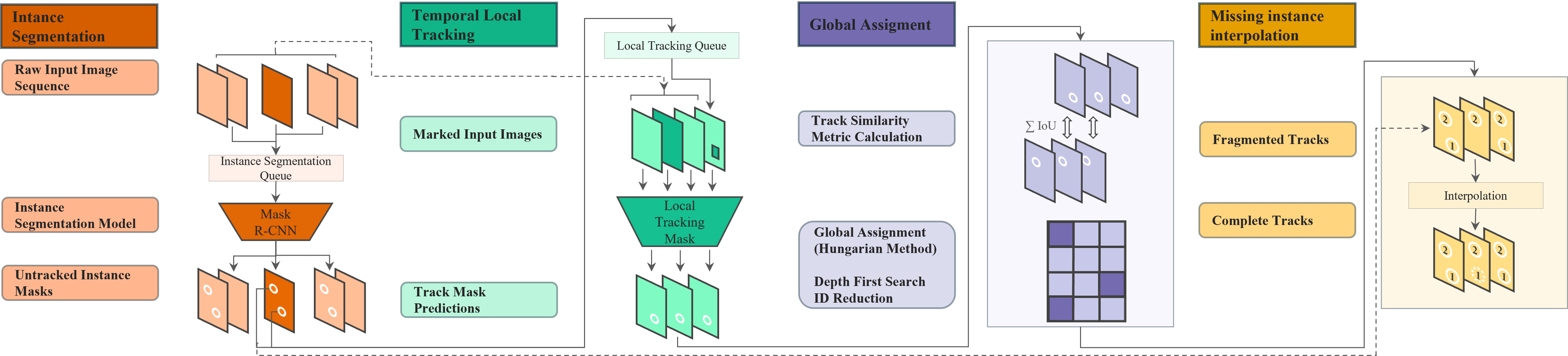}
    \caption{Data flow diagram of the \textit{TS} architecture, illustrating the process from raw input image sequence to finalized track predictions.}
    \label{fig:pipeline_flowchart}
    \vspace{-8pt}
\end{figure*}

The \textit{TS} architecture comprises two primary macro-architectural components: multi-object detection with instance segmentation and multi-object tracking throughout the entire sequence. The tracking component can be further subdivided into temporally local tracking, globally optimal assignment, and missing instance interpolation. While this section provides a brief overview of the architecture, readers seeking a more detailed explanation are encouraged to refer to the original work by Szabó \etal~\cite{szabo2023enhancing}. The high-level data flow of the architecture is illustrated in \cref{fig:pipeline_flowchart}.

Although the performance of the instance segmentation step influences the final results, it employs a fairly standard, approach using a Mask R-CNN \cite{MaskRCNN} architecture with a ResNet-X \cite{ResNet} feature pyramid backbone \cite{FPN}, defined and trained in the Detectron 2 environment \cite{Detectron2}. Therefore, our analysis will primarily focus on the tracking segment of the architecture.

The tracking segment introduces a novel approach by tracking each detected object instance within its local temporal neighborhood, both backward and forward in time, over a range of $TR$ using a semantic segmentation architecture such as DeepLabV3+ \cite{deeplabv3plus2018}. The parameter $TR$ is fixed for a given model, with the $2TR+1$ temporal sequence represented in its input and output channels. An additional input channel serves as a marker for the object being tracked. Metric similarities are then calculated between local track predictions at a maximum temporal offset of $2TR$, allowing for natural re-identification of an object within this range, even if temporarily lost. In practice, the metric used in this scenario is the mean Intersection over Union (IoU) for each matched frame. The globally optimal ID matching between frames is then determined using the Hungarian method \cite{HungarianMethod}, starting with a temporal offset of 1 and continuing to assign missed track endings and starts, up to the maximum offset of $2TR$. This strategy prioritizes higher confidence assignments with lower offsets. After assignment, identical track IDs are organized into a tree graph based on matching identity links, and Depth First Search (DFS) \cite{DFS} is applied to reduce track IDs to the minimal correct number.

\begin{figure}[h!]
    \centering
    \begin{subfigure}[b]{.92\columnwidth}
        \centering
        \includegraphics[width=\linewidth]{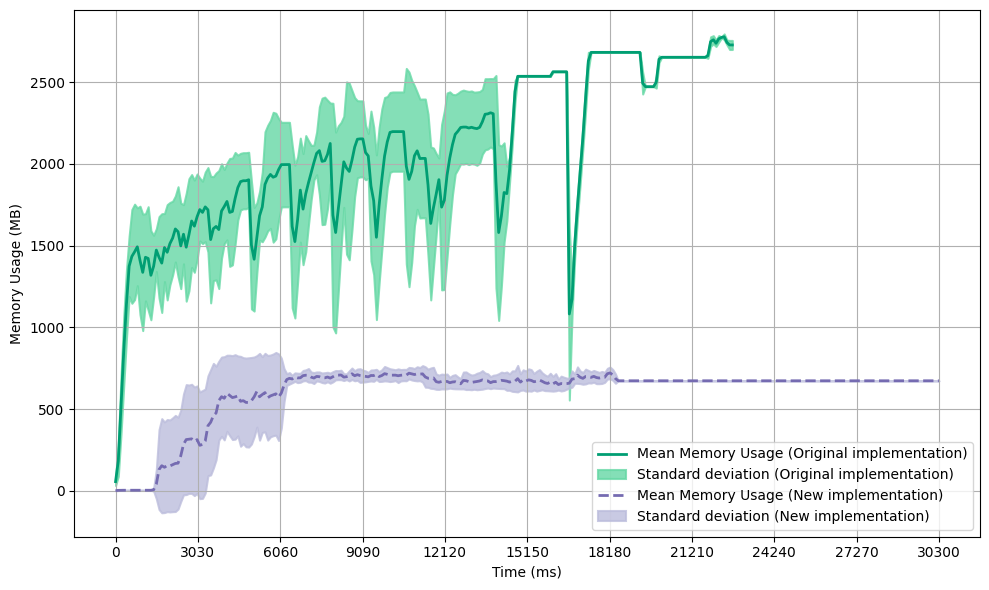}
    \end{subfigure}
    \caption{A display of memory usage and runtime differences between the original implementation of the \textit{TS} architecture tracking segment and our improved implementation.}
    \label{fig:memory_improvement}
    \vspace{-8pt}
\end{figure}

While the original paper by Szabó \etal~\cite{szabo2023enhancing} provides a pipeline implementation, it contains some inefficiencies, resulting in longer-than-necessary inference times and memory overflow errors in longer sequences. Since some of the data we intended to analyze --- specifically the MOTS sequences --- are particularly lengthy, we had to completely refactor and optimize the tracking architecture, while keeping the Detectron 2 \cite{Detectron2} and SMP \cite{SMP} based models intact. The updated pipeline implementation further modularizes the tracking segment by breaking it down into fully independent sub-tasks of data preparation, local tracking, global tracking and graph based ID reduction, employs preparatory calculations to minimize runtime in longer sequences, and mitigates memory usage inflation over time. A comparison of memory usage and runtime between the original and updated \textit{TS} architecture is presented in \cref{fig:memory_improvement}. These results were measured using the five yeast recordings provided through the demonstration environment of the original implementation\cite{szabo2023enhancing}, which are relatively short and do not cause that implementation to crash. While the runtime benefits of the preparatory calculations are not evident in these short sequences, the stark reduction in memory usage inflation in the tracker segment is apparent. The improved version of the full pipeline, frozen at the time of publication, is accessible via the URL provided in the abstract.

\section{Metric definitions} \label{sec:metric_defs}

To evaluate the overall performance of the architecture and specifically highlight details of tracking performance, we employ two families of metrics. The first family is based on comparisons of Intersection over Union (IoU) scores, using a 50\% matching threshold for binary acceptance or rejection, making the interpretation of results straightforward. The second family includes the widely used HOTA scores, along with associated DetA and AssA scores.\cite{HOTA}

\subsection{IoU 50\% scores} \label{sec:metric_defs_IoU50}

The scores in this family are heavily inspired by the F-scores used in the paper by Szabó \etal~\cite{szabo2023enhancing}. However, the tracking F-score used in that paper combines detection and association errors, making it heavily dependent on detection performance. While this approach is valid, we believe that more easily interpretable results can be obtained by focusing solely on association, where detection was successful, akin to the calculation of HOTA. Here, we will define only the association scores, as our primary interest lies in a detailed analysis of tracking performance. For a more general analysis, please refer to the HOTA metric family.

Let $GT(t,n)$ represent the n-th binary ground truth mask at time $t$ and $PD(t,m)$ the m-th binary predicted mask at time $t$. We define the $IoU_{50}$ binary metric as:

{\small
\vspace{-8pt}
\begin{equation}
    IoU(GT(t,n),PD(t,m)) = \frac{|GT(t,n) \cap PD(t,m)|}{|GT(t,n) \cup PD(t,m)|}
\end{equation}
\vspace{-20pt}
\begin{multline}
 IoU_{50}(GT(t,n), PD(t,m)) = \\
\begin{cases} 
 1 & \text{if } IoU(GT(t,n), PD(t,m)) > 0.5 \\
 0 & \text{if } IoU(GT(t,n), PD(t,m)) \leq 0.5
\end{cases}
\end{multline}
\vspace{-8pt}
}

Thus, we can define a successful object detection in a simplified way without needing optimal assignment: if any $GT$ object has a match based on the $IoU_{50}$ metric to any $PD$ object on the same frame, it will be a unique match due to the matching threshold of 50\%, as only a single object can occupy a single image point. Furthermore, if $|GTD_{50}(t,t+1)|$ is the number of ground truth object IDs mutually present at time $t$ and $t+1$ where detection was successful at time $t$, and $|PDD_{50}(t,t+1)|$ is the number of predicted object IDs mutually present at time $t$ and $t+1$ where detection was successful at time $t$, we can define the true positive association count $TPA_{50}$, the false positive association count $FPA_{50}$, and the false negative association count $FNA_{50}$ as follows:
{\small
\begin{align}
 TPA_{50}(t,t+1) 
& = \sum_{n,m} IoU_{50}(GT(t,n), PD(t,m)) \notag \\
&   \cap IoU_{50}(GT(t+1,n), PD(t+1,m))
\end{align}
\begin{equation}
 FPA_{50}(t,t+1) =  |PDD_{50}(t,t+1)| - TPA_{50}(t,t+1)
\end{equation}
\begin{equation}
\textstyle
FNA_{50}(t,t+1) = |GTD_{50}(t,t+1)| - TPA_{50}(t,t+1)
\end{equation}
}
Based on these values, association precision $AP_{50}$, association recall $AR_{50}$, and association F-scores $AF_{50}$ can be calculated for the temporal position $(t,t+1)$ as:
{\small
\begin{equation}
AP_{50} = TPA_{50} / (TPA_{50} + FPA_{50})
\end{equation}
\begin{equation}
AR_{50} = TPA_{50} / (TPA_{50} + FNA_{50})
\end{equation}
\begin{equation}
AF_{50} = \frac{2 AP_{50} AR_{50}}{AP_{50} + AR_{50}}
\end{equation}
}
While precision, recall, and F-scores are typically presented in the range of $[0,1]$, we will present them multiplied by 100 to align with the value range of the HOTA metric family.

\subsection{HOTA metric family}

For a detailed description of the HOTA metric family, please refer to the 2020 paper by Luiten \etal~\cite{HOTA}. In summary, unlike the previously widely used MOTA and IDF1 scores, HOTA effectively balances the importance of both association and detection, making it an excellent metric for evaluating the overall performance of a tracking system. Additionally, HOTA can be decomposed into detection accuracy (DetA) and assignment accuracy (AssA) scores. While our primary focus is on the performance of the tracking model, the overall performance of the entire system is also crucial for interpreting the tracking results. Therefore, DetA, AssA and HOTA scores will be all presented in the analysis. Unlike the segmentation IoU-based metrics defined in \cref{sec:metric_defs_IoU50}, these scores are computed for the bounding boxes of each object instance to align better with comparisons to other tracking models that typically predict only bounding boxes, not segmentations. Hence, we advise interpreting the IoU 50\% scores as the primary indicators of model performance, with the HOTA metric family serving as a comparative measure that may obscure some of the true potential of the segmentation-based tracker.

\section{Tracker variants}\label{sec:tracker_variants}

To evaluate the tracking capabilities of the \textit{TS} architecture, we compared it with the widely used Kalman filter \cite{KalmanFilter}, a default choice in many object tracking environments. Although advancements have made the mathematical limitations regarding linearity and parameterization of the Kalman filter more flexible \cite{UnscentedKalman,NeuralKalman}, the core concept remains the same: tracking objects based on a limited set of parameters such as position and its derivatives, while the Kalman filter optimally balances the estimated temporal forward prediction between past positions, updated according to a state transition model, and new object detections, transformed by an observation model. Although this temporal forward prediction approach is computationally lightweight, making it ideal for live object tracking scenarios, it disregards morphological information and other input data details not represented by the observation model. Moreover, temporal forward predictions must be assigned to observations at different temporal positions to perform tracking, which involves assigning objects in different state spaces. While this might seem sound at first glance, the similarity measurement is highly constrained by the state transition and observation models, limiting the information that can be compared between observed and forward-predicted objects. This often defaults to an oversimplified metric, such as L2 distance between centroid positions. Additionally, any temporal forward prediction model ignores future data, which is reasonable for live tracking scenarios but overlooks potential information when the entire temporal sequence is available at the time of prediction.

In contrast, the \textit{TS} architecture's local tracking segment inherently learns temporally local behaviors and predicts segmented masks using both past and future data. These predictions are compared in the same state space, as they are made by the same model from different perspectives. We believe that the difference between temporally forward-predicting models and the symmetric tracking capabilities of the \textit{TS} architecture is somewhat analogous to a comparison between the Forward algorithm \cite{ForwardAlgorithm} and the Viterbi algorithm \cite{ViterbiAlgorithm}, although the temporally symmetric parallel predictions and the updating of past predictions based on new information in the Viterbi algorithm are conceptually distinct. On the other hand, while it is natural for the \textit{TS} architecture to skip missed object instances and re-interpolate them after tracking, temporal forward predictors like the Kalman filter can also skip missed instances if the state transition model is applied multiple times after a matching detection is not found based on the matching criterion. In our implementation of the Kalman filter-based tracker, we allowed temporal forward predictions and re-interpolations with a maximum of 8 frames distance, matching the local tracker models' $TR$ value of 4, resulting in a maximum assignment distance of $2TR = 8$.

To further assess the impact of positional and morphological information in assignments, we evaluated two restricted variants of the \textit{TS} architecture: \textit{TS-L2} and \textit{TS-Shape}. The \textit{TS-L2} variant uses the same local tracking model but retains only the centroids of the predicted masks for L2 distance-based similarity comparison, ignoring all morphological information. The \textit{TS-Shape} variant aligns the centroids of predicted masks before calculating IoU-based similarity, focusing solely on morphology and disregarding positional data. Evaluating the \textit{TS-L2} variant is particularly interesting, as it serves as a middle ground between the Kalman filter and the \textit{TS} architecture by using visual cues from inputs while ignoring morphology during assignment.

Despite the substantial differences in prediction methodologies among these four models, their predictions can be handled similarly. We applied the unaltered Hungarian algorithm-based global assignment step of the \textit{TS} architecture, followed by depth-first search of connected IDs and interpolation of missed instances to all predictions. Furthermore it must be noted that following track prediction, we omitted any tracks shorter than 10 frames in length. While this introduces a reverse dependence from tracking quality to detection quality, it also adds a sense of realism to the tests, as in autonomous systems such low confidence detections are often omitted too.
 
\section{Datasets} \label{sec:datasets}

While evaluating novel methods on natural datasets is crucial, such datasets for MOTS (Multi-Object Tracking and Segmentation) tasks are relatively rare, especially those with ample training data where all objects are accurately labeled and tracked. For instance, the recently released SA-V dataset, allegedly used to train the SAM 2 model, provides a large number of high-quality tracks but only for a few objects per recording \cite{SAM2}. This limitation makes it unsuitable for training the instance segmentation stage of the evaluated architectures and almost completely prevents the assessment of the recall capabilities. Moreover, the \textit{TS} architecture has already been proven to outperform other methods that were specifically designed for budding yeast cells dataset of such videomicroscopic recordings.

Therefore, we opted to create various synthetic scenarios, strongly inspired by the synthetic datasets presented in the original publication of the \textit{TS} architecture \cite{szabo2023enhancing}, to evaluate specific performance differences among the four tracking models described in \cref{sec:tracker_variants}. The code used to generate these scenarios, along with the resulting datasets, is available at the URL provided in the abstract. Although these scenarios are artificial, their aim is to simulate key features and events commonly observed in natural settings. Additionally, to ensure evaluation on natural datasets, we trained models on the synthetic MOTSynth-MOTS-CVPR22 dataset \cite{MOTSynth} and then evaluated their performance on real-world samples from the MOTS dataset \cite{MOTS}.

\subsection{Visual signaling scenario} \label{sec:visual_signaling}

For the foundation of this synthetic dataset group, we used the \textit{Synthetic Arrows} scenario presented in the publication of the \textit{TS} architecture \cite{szabo2023enhancing}. This scenario was selected for its simplicity: the objects move quickly with near-linear motion characteristics, and there is minimal visual information to be gained from the objects' morphological features, except for the indication of their forward direction. Consequently, we anticipated the Kalman filter to perform well in this scenario, providing a benchmark for evaluating the \textit{TS} architecture variants. In contrast, the modified versions of the \textit{Synthetic Arrows} dataset that we created were expected to be more challenging. In these scenarios, the arrows undergo a specific color change before a turning event, similar to how cars use turn signals, indicating the direction in which the arrow will turn after a number of $T$ frames. The arrow then executes a 90\textdegree turn in the signaled direction. The first variant, \textit{Synthetic Arrows TR-1}, initiates a turn with a 20\% chance per frame, with a signaling period of $T = 4$ frames. The second variant, \textit{Synthetic Arrows TR-2}, initiates a turn with an 80\% chance per frame, with a signaling period of $T = 2$ frames. These scenarios were designed to test how well the \textit{TS} variants interpret visual cues and to assess the extent to which visually signaled kinematic events disrupt the Kalman filter.

\subsection{Semi-random positioning scenario} \label{sec:random_positioning}

Similarly to the "Visual Signaling" scenario, we also used the \textit{Synthetic Amoeboids} scenario from the \textit{TS} architecture publication \cite{szabo2023enhancing} as a baseline. However, we applied no morphological changes to the objects from one frame to the next, making them even more recognizable based on their morphological features. Unlike the \textit{Synthetic Arrows} scenario, the amoeboids possess unique morphological characteristics due to the Perlin noise \cite{PerlinNoise} applied in their generation process. While these objects are still relatively easy to track, we anticipated that disregarding morphological cues would be disadvantageous for both the Kalman filter and the \textit{TS-L2} model, even for this baseline scenario, especially when compared to \textit{Synthetic Arrows}. The modified versions of the \textit{Synthetic Amoeboids} baseline dataset that we created were also expected to be increasingly challenging. While in these variants, the objects still exhibit semi-random but almost linear movement patterns, excluding collisions, their final position in each frame is adjusted by a random uniform positional noise at a maximum distance of $1/D$ relative to the field of view in both the x and y directions. As $1/D$ increases, this makes object tracking solely based on position progressively more difficult, thereby raising the importance of object morphology. The two variants of this scenario are \textit{Synthetic Amoeboids RP-1/20} and \textit{Synthetic Amoeboids RP-1/5}, with random positioning relative distances of 1/20 and 1/5, respectively. We expect these variants to be particularly challenging for tracking models that ignore morphological information.

\subsection{MOTS challenge} \label{sec:MOTS}

The MOTSynth-MOTS-CVPR22 training dataset includes 767 full HD videos, each 1,800 frames long, generated within the computer game GTA V \cite{MOTSynth}, with pedestrians annotated as objects. The official test set consists of seven naturally captured and manually annotated samples. However, because the challenge’s evaluation kit has been outdated for some time, we opted instead to use three publicly released full HD resolution sequences from the MOTS challenge training set as our test set, available at \url{https://motchallenge.net/data/MOTS/} (2024). Our models were trained on the first 600 recordings of the training set, with the remaining training samples used for evaluation. Notably, we were unable to perform deep fine-tuning of model hyperparameters on the training dataset either, due to the immense computational requirements and training times. Thus, we opted to use the original augmentation scheme and hyperparameters of the \textit{TS} model, with a local tracking range of 2. Therefore, it is likely that, with a more specialized model, even better results could be achieved.

To the best of our knowledge, the only official submission to the MOTSynth-MOTS-CVPR22 challenge is based on the widely recognized "Tracking without Bells and Whistles" (\textit{Tracktor}) model by Bergmann \etal~\cite{TracktorPP}. Although only this single submission exists, it employs a popular state-of-the-art tracking method, making its performance the most reliable benchmark for comparison in our evaluation.

\section{Evaluation results}

We conducted a comprehensive evaluation of the tracker models \textit{Kalman} filter, \textit{TS}, \textit{TS-L2}, and \textit{TS-Shape} using the metrics defined in \cref{sec:metric_defs} across both synthetic scenarios and the MOTS challenge variant outlined in \cref{sec:datasets}. Additionally, we performed an attention analysis of the local tracker model within the \textit{TS} architecture, the details of which are presented later in \cref{sec:saliency}. The evaluation results are presented as Kernel Density Estimates (KDE) \cite{KDE} for the AssA and HOTA metrics, and as mean values for all other metrics. For proper interpretation of the results, it is important to note that an observable increase in DetA and HOTA scores often coincides with particularly low AssA and especially $AR_{50}$ scores. While these elevated DetA and HOTA values are technically correct, they primarily result from the exclusion of unreasonably short tracks, which removes lower-confidence detection instances and boosts detection precision, thereby inflating DetA and HOTA scores. Therefore, as our focus is primarily on tracking performance, we caution against drawing far-reaching conclusions from these inflated DetA and HOTA values.

\subsection{Synthetic scenarios}

\begin{figure}[h!]
    \centering
    \begin{subfigure}[b]{.92\columnwidth}
        \centering
        \includegraphics[width=\linewidth]{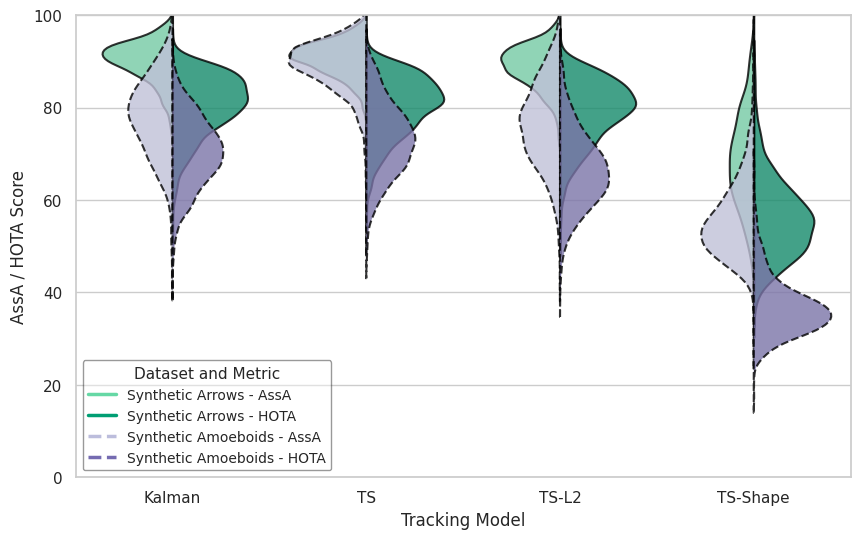}
    \end{subfigure}
     \begin{subfigure}[b]{.92\columnwidth}
        \centering
        \includegraphics[width=.66\linewidth]{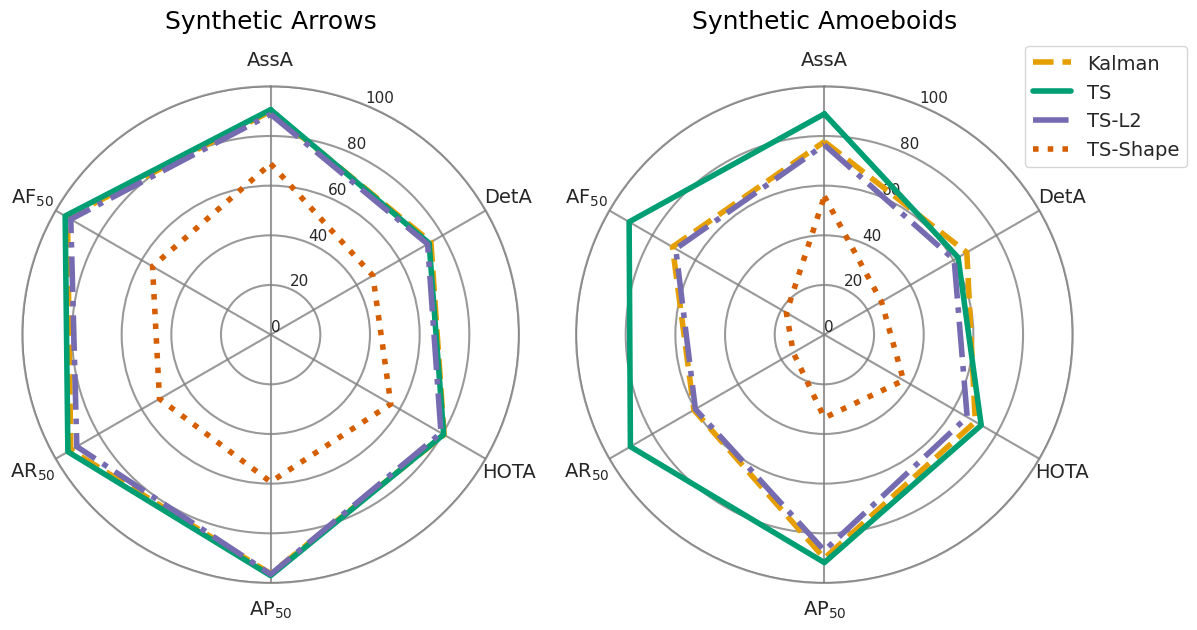}
    \end{subfigure}
    \caption{KDE (top) and mean (bottom) metric results of tracker models \textit{Kalman}, \textit{TS}, \textit{TS-L2} and \textit{TS-Shape} for datasets \textit{Synthetic Arrows} and \textit{Synthetic Amoeboids}.}
    \label{fig:SarrSamb}
    \vspace{-8pt}
\end{figure}

To establish a baseline for the other scenarios, we first present the comparative evaluation results between the \textit{Synthetic Arrows} and \textit{Synthetic Amoeboids} datasets in \cref{fig:SarrSamb}. While the models \textit{Kalman}, \textit{TS}, and \textit{TS-L2} demonstrated similar performance in the simplistic \textit{Synthetic Arrows} scenario, the \textit{TS} architecture clearly outperformed both the Kalman filter and the \textit{TS-L2} model in the morphologically more complex \textit{Synthetic Amoeboids} scenario. This result highlights the potential benefit of incorporating morphological information. However, the \textit{TS-Shape} model substantially underperformed on both datasets, particularly in the \textit{Synthetic Amoeboids} scenario, indicating that positional information alone is more valuable in these scenarios than morphological information, and that the combined benefit of both types of information is not merely additive. Notably, the better performance of the \textit{TS-Shape} model on the \textit{Synthetic Arrows} is likely due to the arrows being often accurately identifiable by their specific area, whereas different amoeboids might share similar surface features by chance.

\begin{figure}[h!]
    \centering
    \begin{subfigure}[b]{.92\columnwidth}
        \centering
        \includegraphics[width=\linewidth]{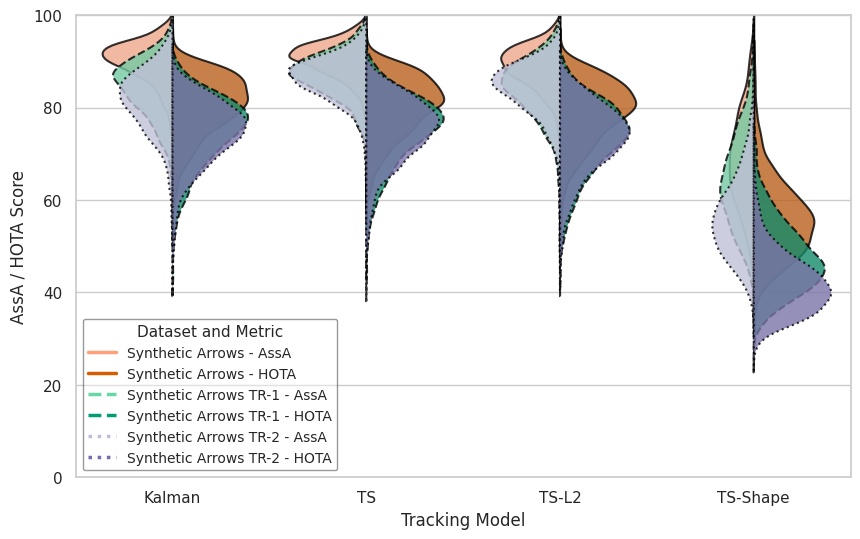}
    \end{subfigure}
     \begin{subfigure}[b]{.92\columnwidth}
        \centering
        \includegraphics[width=\linewidth]{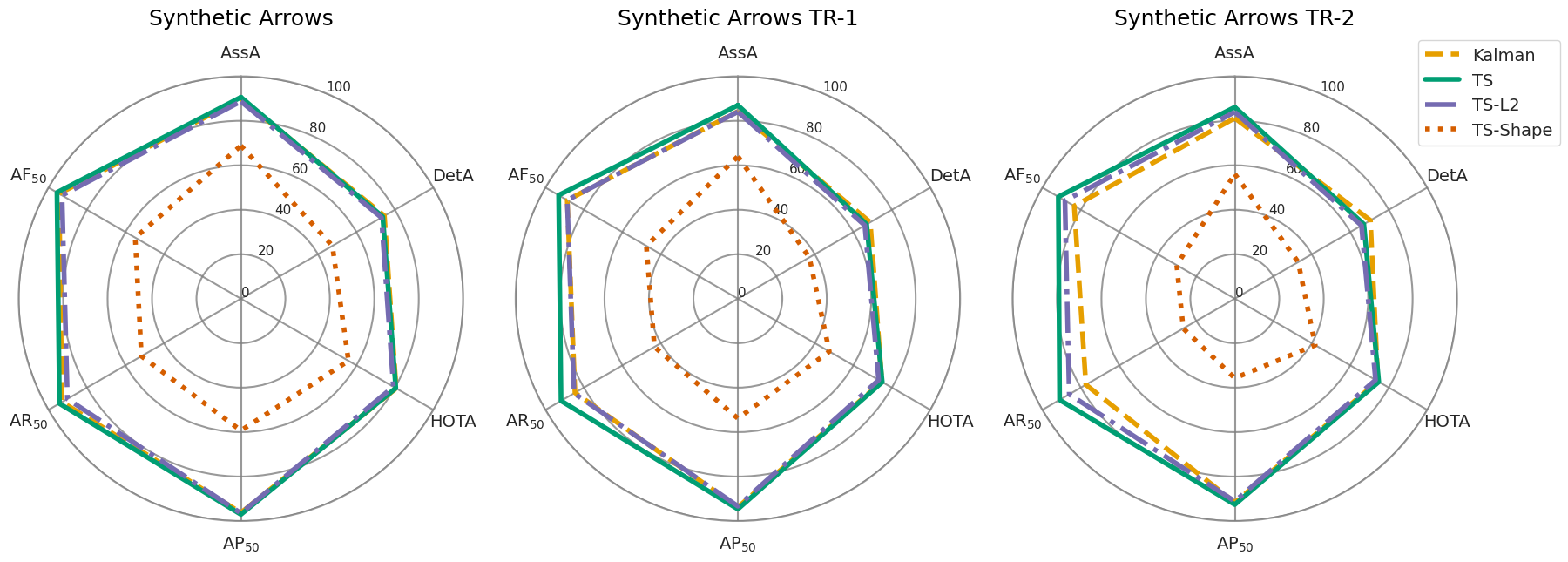}
    \end{subfigure}
    \caption{KDE (top) and mean (bottom) metric results of tracker models \textit{Kalman}, \textit{TS} and \textit{TS-L2}, \textit{TS-Shape} for scenario "Visual signaling" datasets \textit{Synthetic Arrows}, \textit{Synthetic Arrows TR-1} and \textit{Synthetic Arrows TR-2}.}
    \label{fig:SArr}
    \vspace{-8pt}
\end{figure}

Next, we present the results for the "Visual signaling" scenario defined in \cref{sec:visual_signaling} in \cref{fig:SArr}. As anticipated, the Kalman filter's performance deteriorates as the movement patterns of the objects become more dependent on visual signals. In contrast, the \textit{TS} and \textit{TS-L2} architectures show a lesser decline in performance, with the difference being clearly noticeable but relatively modest. Notably, the \textit{TS-L2} architecture performs similarly to the \textit{TS} architecture, as the visual signals are encoded within the architecture and positional estimates, even though the assignment metric in the architecture disregards morphological information. The \textit{TS-Shape} architecture continues to perform the worst, as there is minimal morphological information available to differentiate the objects.

\begin{figure}[h!]
    \centering
    \begin{subfigure}[b]{.92\columnwidth}
        \centering
        \includegraphics[width=\linewidth]{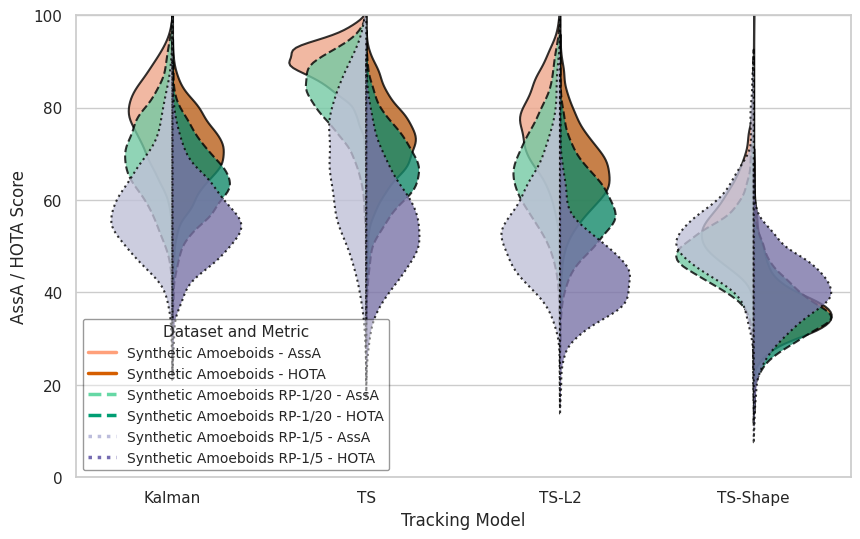}
    \end{subfigure}
     \begin{subfigure}[b]{.92\columnwidth}
        \centering
        \includegraphics[width=\linewidth]{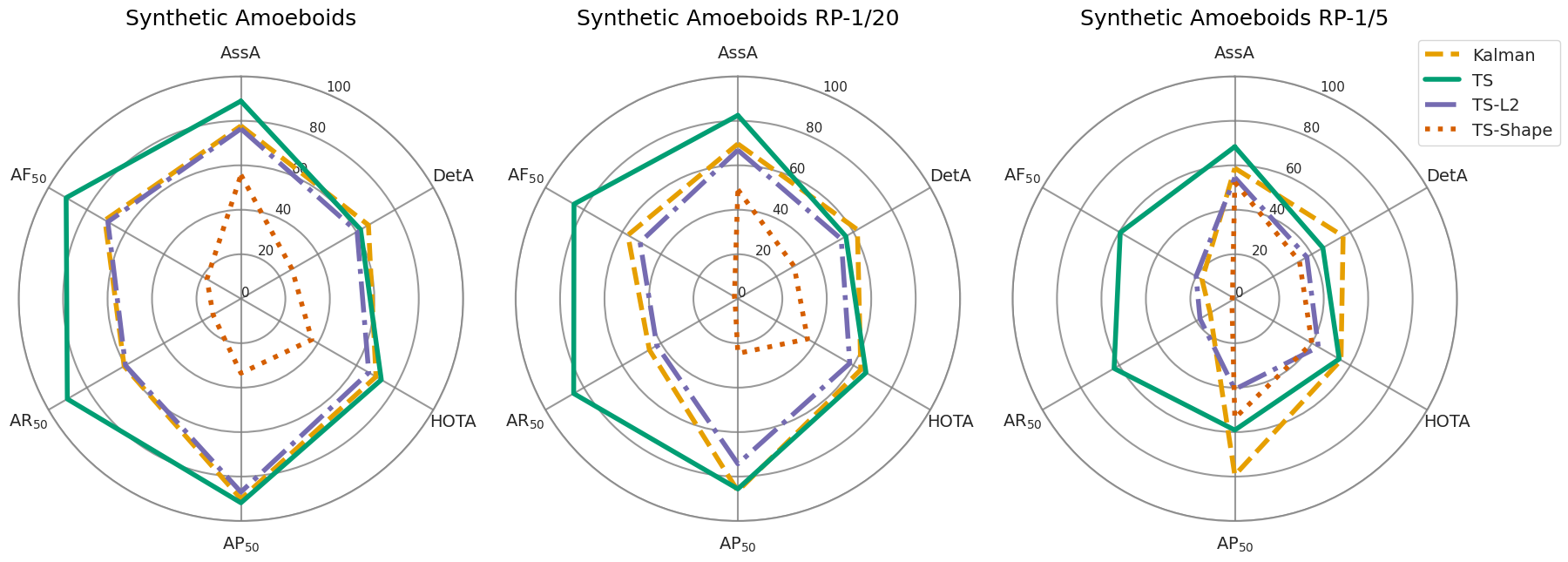}
    \end{subfigure}
    \caption{KDE (top) and mean (bottom) metric results of tracker models \textit{Kalman}, \textit{TS}, \textit{TS-L2} and \textit{TS-Shape} for scenario "Semi-random positioning" datasets \textit{Synthetic Amoeboids}, \textit{Synthetic Amoeboids RP-1/20} and \textit{Synthetic Amoeboids RP-1/5}.}
    \label{fig:SAmb}
    \vspace{-8pt}
\end{figure}

Lastly, we present the results for the "Semi-Random Positioning" scenario defined in \cref{sec:random_positioning} in \cref{fig:SAmb}. Here, the \textit{TS} architecture shows a clear advantage over both the Kalman filter and the \textit{TS-L2} architecture, emphasizing the benefit of predicting full object morphologies instead of relying solely on positional assignments. While the \textit{TS-Shape} architecture still performs far worse than the others, its performance remains relatively stable across the scenario, as it is unaffected by the position of the objects.

\subsection{MOTS challenge}

\begin{figure}[h!]
    \centering
    \begin{subfigure}[b]{.92\columnwidth}
        \centering
        \includegraphics[width=\linewidth]{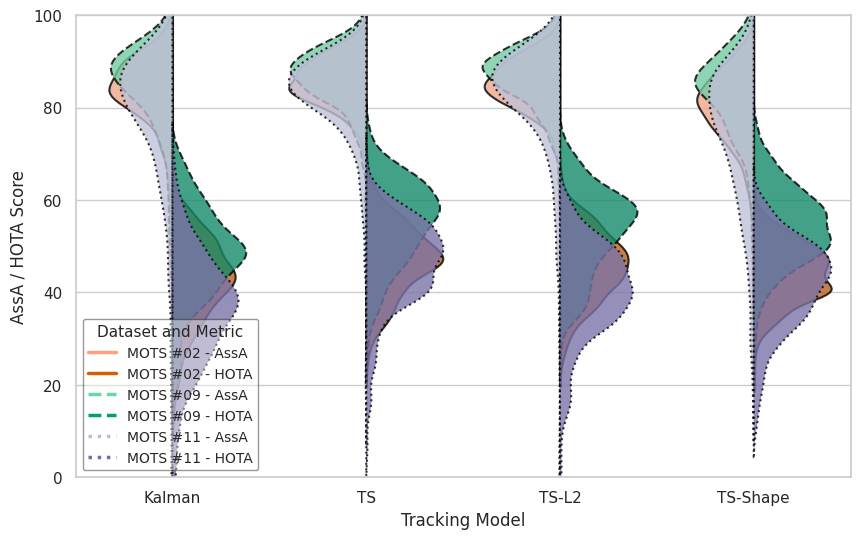}
    \end{subfigure}
     \begin{subfigure}[b]{.92\columnwidth}
        \centering
        \includegraphics[width=\linewidth]{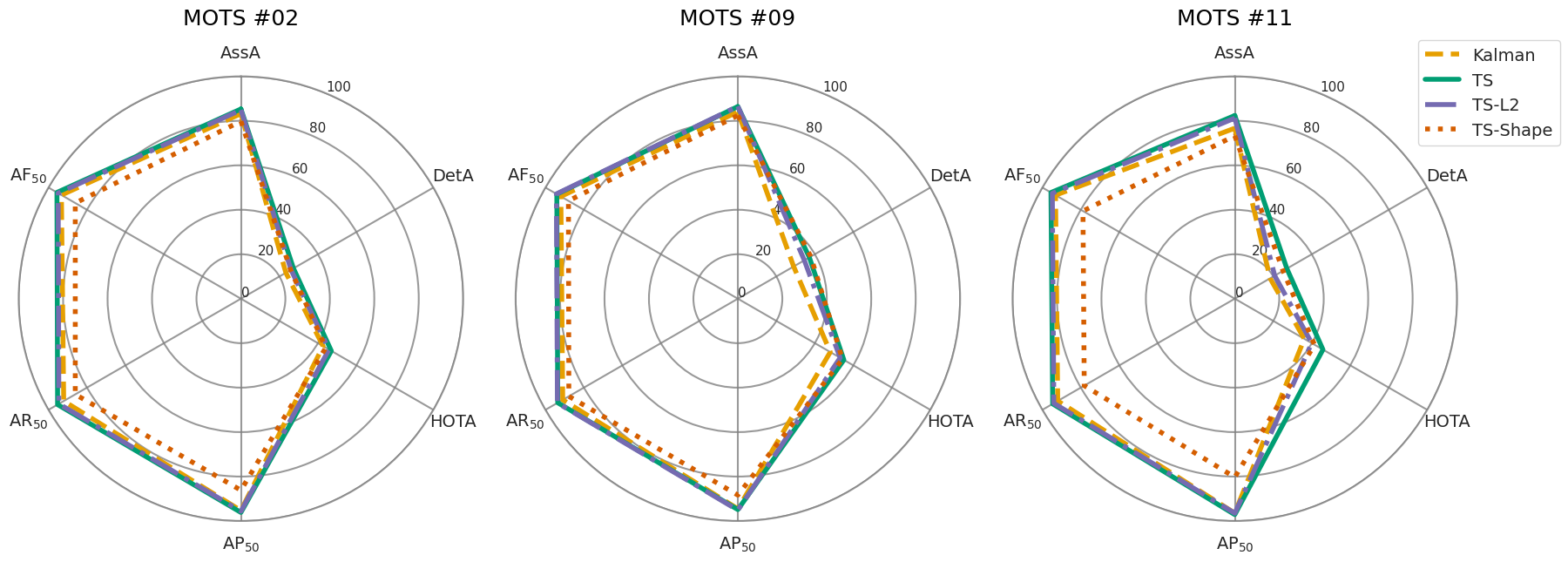}
    \end{subfigure}
    \caption{KDE (top) and mean (bottom) metric results of tracker models \textit{Kalman}, \textit{TS}, \textit{TS-L2} and \textit{TS-Shape} for the MOTS dataset samples described in \cref{sec:MOTS}}
    \label{fig:MOTS}
    \vspace{-8pt}
\end{figure}

Similarly to the synthetic scenarios, we present the results for the MOTS challenge samples described in \cref{sec:MOTS} in \cref{fig:MOTS}. Due to the particularly challenging nature of this task, the object detection score is low. Still, the \textit{TS} model performs notably better than the other models. Interestingly, while the \textit{TS-Shape} architecture still has the lowest performance overall, it performs much better than in the synthetic scenarios. We believe this can be attributed partially to the particularly clear visual differences between the pedestrian objects to be tracked.

The overall performance of the \textit{TS} architecture, with a mean HOTA score of $48.56$, closely matches the HOTA score of $48.8$ achieved by the benchmark \textit{Tracktor} model. Moreover, the \textit{TS} architecture achieves a substantially higher mean AssA score of $82.39$, compared to the \textit{Tracktor} model’s AssA score of $44.6$. This suggests that the comparable HOTA scores are primarily due to imperfect detection and segmentation by the Detectron 2 based Mask R-CNN instance segmentation step within the \textit{TS} architecture, while its novel tracking approach far outperforms the \textit{Tracktor} model. Furthermore, since the Mask R-CNN with a ResNet-X feature pyramid backbone used by the \textit{TS} architecture is widely recognized as one of the top-performing instance segmentation models \cite{murthy2020investigations, abdusalomov2023improved, pham2020road}, it is highly likely that with a more specialized training scheme of the instance segmentation model —-- including dataset-specific data augmentation and hyperparameter tuning —-- a substantially higher HOTA score could be achieved.

\subsection{Attention analysis} \label{sec:saliency}

In addition to the various comparative performance analyses conducted, we believe that examining the attention of the local tracker model within the \textit{TS} architecture can provide further key insights into its inner workings. We performed this analysis on the \textit{Synthetic Arrows} dataset using both a local tracking model pretrained on the ImageNet dataset \cite{ImageNet} and a similar model trained for the same number of steps but with randomly initialized weights. The models showed comparable practical results, with the pretrained model achieving an AssA value of $90.83$ and a HOTA value of $81.24$, while the non-pretrained model achieved an AssA value of $90.25$ and a HOTA value of $80.63$, with the pretrained model performing marginally better.

Attention or saliency maps can be generated in various ways, even for single-output classification architectures \cite{SaliencyMapsBase,SaliencyMapsSanity,SaliencyMapsBálint}. For our analysis, we employed a relatively simple method to calculate attention maps by backpropagating the prediction value of the predicted mask's centroid after the prediction of the multi-channel mask, either for the central frame alone or across all prediction channels. For the latter, the attention maps for each input channel were averaged across all prediction channels. However, as both spatial and temporal distributions of attention were near identical between the two approaches, we only present the results based on the average across all channels. While this method does not capture the attention over the entire segmented area, it provides a representative result across many observed objects, and it is computationally far less demanding. We performed the calculations on all objects over ten frames from a single test sample, as the mean spatial and temporal distribution of the attention maps quickly converged to a stable state, yielding practically indistinguishable results across multiple runs.

\begin{figure}[h!]
    \centering
    \begin{subfigure}[b]{.87\columnwidth}
        \centering
        \includegraphics[width=\linewidth]{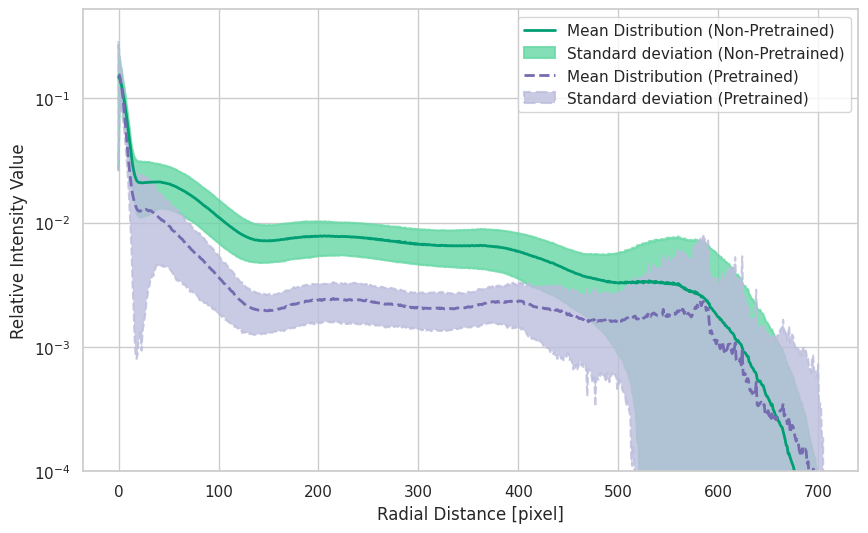}
    \end{subfigure}
     \begin{subfigure}[b]{.87\columnwidth}
        \centering
        \includegraphics[width=\linewidth]{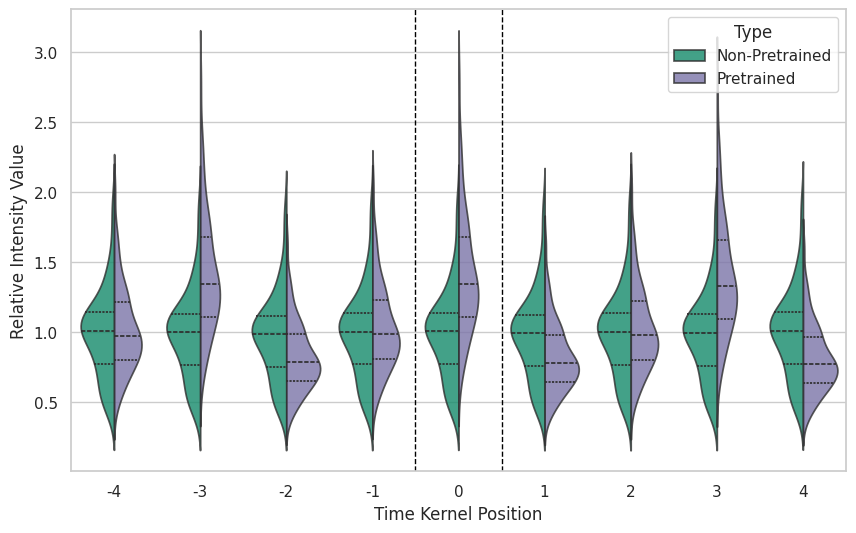}
    \end{subfigure}
    \caption{A display of radial (top) and temporal (bottom) relative distribution of attention for the local tracking model of the \textit{TS} architecture, trained on the \textit{Synthetic Arrows} dataset, based on the backpropagation of the predicted value for the centroid of the object to be tracked.}
    \label{fig:saliency}
    \vspace{-12pt}
\end{figure}

To provide a comprehensive numerical analysis of the attention mechanisms in the local tracking model, we measured the relative distribution of attention values as a function of radial distance from the object's centroid in the central frame and as a function of the temporal channel positions within the model's input. The results of this analysis are presented in \cref{fig:saliency}. These findings indicate that both the pretrained and non-pretrained models exhibit a fairly centralized radial attention distribution, as expected. However, the non-pretrained model places greater emphasis on less localized features. Notably, the pretrained model shows a distinct repetitive pattern in channel preference, likely inherited from the RGB color channel biases of the original ImageNet-trained model. In contrast, the non-pretrained model demonstrates a uniform distribution, which aligns with expectations. From a theoretical standpoint, these results suggest that avoiding pretraining with a transfer of color channels to temporal channels may be preferable. However, the practical results reveal no substantial difference between the models. Furthermore, in more complex image processing scenarios, the practical advantages of using pretrained models might become more significant.

\section{Discussion}

Even from the baseline comparison of the \textit{Synthetic Arrows} and \textit{Synthetic Amoeboids} datasets, the clear advantage of the \textit{TS} tracking architecture over the Kalman filter is evident. The \textit{TS} architecture’s ability to process multimodal information rather than just kinematic data, proves beneficial when such information is relevant. Furthermore, the results suggest that the optimal integration of positional and morphological information for track assignment far outperforms the use of either information alone.

In the "Visual Signaling" scenario, the results reveal that incorporating movement pattern forecasting based on visual signals can enhance movement prediction, even when morphological information is not used in the track assignment step. This is demonstrated by the increasingly superior performance of the \textit{TS-L2} model compared to the Kalman filter as the frequency of color-signaled motility events increases. This highlights the importance of visual cues in multi-object tracking, such as signal lights, turn signals, head movements, and even subtle gait changes often observed in real-world environments. Notably, while human perception and real-time tracking models can only utilize past and present data, it is possible --- and likely --- that visual cues appearing in the future can aid in predicting motion that occurred before those cues. Processing these cues at inference time can be advantageous, and the \textit{TS} architecture is uniquely capable of capturing and utilizing this information.

In the "Semi-random Positioning" scenario, where object movement is less predictable based on past positions and morphological information becomes crucial, the \textit{TS} architecture shows clear superiority over the Kalman filter. Furthermore, in the \textit{Synthetic Amoeboids RP-1/5} dataset, the performances of the \textit{TS-L2} and \textit{TS-Shape} models nearly balance out, demonstrating the increased importance of morphological information when positional estimates are unstable or when lower temporal sampling rates lead to less consistent movement patterns.

On the MOTS dataset, the \textit{TS} architecture achieved a HOTA score comparable to that of the benchmark \textit{Tracktor} model, reflecting similar overall performance. However, the associative tracking performance of the \textit{TS} architecture far surpassed the \textit{Tracktor} benchmark. This strongly suggests that the \textit{TS} architecture offers superior tracking capabilities, and with more dataset-specific training and optimization of the instance segmentation component within the \textit{TS} architecture, substantially higher overall tracking results could likely be achieved on the MOTSynth-MOTS-CVPR22 dataset. Moreover, this performance highlights the \textit{TS} architecture's potential applicability beyond videomicroscopic cell tracking to offline personnel tracking tasks such as surveillance, crowd movement analysis, and other related applications. Additionally, the \textit{TS} tracking architecture's ability to deliver state-of-the-art performance across vastly different tasks strongly indicates its suitability for a wide range of tracking applications, provided its lower inference speed is acceptable in exchange for predictive performance.

The attention analysis of the local tracking model shows that when pretrained on colored images, the architecture may retain initial channel attentions, leading to a potentially flawed temporal attention pattern. However, the pretrained model performed marginally better with a 0.58 AssA score difference. This suggests that either the advantages and drawbacks of pretraining balance out, or that the attention differences do not meaningfully impact performance.

\section{Conclusion}

Using multiple synthetic scenarios and the MOTS challenge dataset, we conducted an in-depth analysis of the instance segmentation and tracking architecture, with a specific focus on the novel tracking model (\textit{TS}) introduced by Szabó \etal~\cite{szabo2023enhancing}. Although the architecture was originally developed for tracking budding yeast cells, where it demonstrated outstanding performance, our evaluation scenarios were designed to highlight specific modalities not necessarily present in that environment. We performed detailed comparisons against two restricted variants of the local tracking model and the widely used Kalman filter. The results clearly demonstrate the architectural advantages of the \textit{TS} model and highlight the limitations of the alternative approaches. These findings emphasize the importance of selecting the appropriate architecture when dealing with potential morphological and other visual cues. Additionally, we improved the original implementation of the \textit{TS} architecture, achieving state-of-the-art overall segmentation and tracking performance on the MOTS personnel tracking dataset, while delivering far superior associative tracking performance compared to the benchmark \textit{Tracktor} model. Based on these findings, while there is room for further refinement of the \textit{TS} architecture, we believe that its unique approach and demonstrated capabilities make it a flexible, state-of-the-art solution for simultaneous instance segmentation and tracking on pre-recorded data.


{
    \small
    \bibliographystyle{ieeenat_fullname}
    \bibliography{main}

\begin{thebibliography}{28}
\providecommand{\natexlab}[1]{#1}
\providecommand{\url}[1]{\texttt{#1}}
\expandafter\ifx\csname urlstyle\endcsname\relax
  \providecommand{\doi}[1]{doi: #1}\else
  \providecommand{\doi}{doi: \begingroup \urlstyle{rm}\Url}\fi

\bibitem[Abdusalomov et~al.(2023)Abdusalomov, Islam, Nasimov, Mukhiddinov, and Whangbo]{abdusalomov2023improved}
Akmalbek~Bobomirzaevich Abdusalomov, Bappy MD~Siful Islam, Rashid Nasimov, Mukhriddin Mukhiddinov, and Taeg~Keun Whangbo.
\newblock An improved forest fire detection method based on the detectron2 model and a deep learning approach.
\newblock \emph{Sensors}, 23\penalty0 (3):\penalty0 1512, 2023.

\bibitem[Adebayo et~al.(2018)Adebayo, Gilmer, Muelly, Goodfellow, Hardt, and Kim]{SaliencyMapsSanity}
Julius Adebayo, Justin Gilmer, Michael Muelly, Ian Goodfellow, Moritz Hardt, and Been Kim.
\newblock Sanity checks for saliency maps.
\newblock \emph{Advances in neural information processing systems}, 31, 2018.

\bibitem[Al-Afandi et~al.(2022)Al-Afandi, Magyar, and Horv{\'a}th]{SaliencyMapsBálint}
Jalal Al-Afandi, B{\'a}lint Magyar, and Andr{\'a}s Horv{\'a}th.
\newblock Saliency map based data augmentation.
\newblock In \emph{2022 26th International Conference on Pattern Recognition (ICPR)}, pages 4751--4757. IEEE, 2022.

\bibitem[Baum and Petrie(1966)]{ForwardAlgorithm}
Leonard~E Baum and Ted Petrie.
\newblock Statistical inference for probabilistic functions of finite state markov chains.
\newblock \emph{The annals of mathematical statistics}, 37\penalty0 (6):\penalty0 1554--1563, 1966.

\bibitem[Bergmann et~al.(2019)Bergmann, Meinhardt, and Leal-Taixe]{TracktorPP}
Philipp Bergmann, Tim Meinhardt, and Laura Leal-Taixe.
\newblock Tracking without bells and whistles.
\newblock In \emph{Proceedings of the IEEE/CVF international conference on computer vision}, pages 941--951, 2019.

\bibitem[Chen et~al.(2018)Chen, Zhu, Papandreou, Schroff, and Adam]{deeplabv3plus2018}
Liang-Chieh Chen, Yukun Zhu, George Papandreou, Florian Schroff, and Hartwig Adam.
\newblock Encoder-decoder with atrous separable convolution for semantic image segmentation.
\newblock In \emph{ECCV}, 2018.

\bibitem[Deng et~al.(2009)Deng, Dong, Socher, Li, Li, and Fei-Fei]{ImageNet}
Jia Deng, Wei Dong, Richard Socher, Li-Jia Li, Kai Li, and Li Fei-Fei.
\newblock Imagenet: A large-scale hierarchical image database.
\newblock In \emph{2009 IEEE Conference on Computer Vision and Pattern Recognition}, pages 248--255, 2009.

\bibitem[Fabbri et~al.(2021)Fabbri, Bras{\'o}, Maugeri, Cetintas, Gasparini, O{\v{s}}ep, Calderara, Leal-Taix{\'e}, and Cucchiara]{MOTSynth}
Matteo Fabbri, Guillem Bras{\'o}, Gianluca Maugeri, Orcun Cetintas, Riccardo Gasparini, Aljo{\v{s}}a O{\v{s}}ep, Simone Calderara, Laura Leal-Taix{\'e}, and Rita Cucchiara.
\newblock Motsynth: How can synthetic data help pedestrian detection and tracking?
\newblock In \emph{Proceedings of the IEEE/CVF International Conference on Computer Vision}, pages 10849--10859, 2021.

\bibitem[Forney(1973)]{ViterbiAlgorithm}
G.D. Forney.
\newblock The viterbi algorithm.
\newblock \emph{Proceedings of the IEEE}, 61\penalty0 (3):\penalty0 268--278, 1973.

\bibitem[He et~al.(2016)He, Zhang, Ren, and Sun]{ResNet}
Kaiming He, Xiangyu Zhang, Shaoqing Ren, and Jian Sun.
\newblock Deep residual learning for image recognition.
\newblock In \emph{Proceedings of the IEEE conference on computer vision and pattern recognition}, pages 770--778, 2016.

\bibitem[He et~al.(2017)He, Gkioxari, Doll{\'a}r, and Girshick]{MaskRCNN}
Kaiming He, Georgia Gkioxari, Piotr Doll{\'a}r, and Ross Girshick.
\newblock Mask r-cnn.
\newblock In \emph{Proceedings of the IEEE international conference on computer vision}, pages 2961--2969, 2017.

\bibitem[Iakubovskii(2019)]{SMP}
Pavel Iakubovskii.
\newblock Segmentation models pytorch.
\newblock \url{https://github.com/qubvel/segmentation_models.pytorch}, 2019.

\bibitem[Kuhn(1955)]{HungarianMethod}
Harold~W Kuhn.
\newblock The hungarian method for the assignment problem.
\newblock \emph{Naval research logistics quarterly}, 2\penalty0 (1-2):\penalty0 83--97, 1955.

\bibitem[Kálmán(1960)]{KalmanFilter}
Rudolf~E. Kálmán.
\newblock A new approach to linear filtering and prediction problems.
\newblock \emph{Journal of Basic Engineering}, 82\penalty0 (1):\penalty0 35--45, 1960.

\bibitem[Lin et~al.(2017)Lin, Doll{\'a}r, Girshick, He, Hariharan, and Belongie]{FPN}
Tsung-Yi Lin, Piotr Doll{\'a}r, Ross Girshick, Kaiming He, Bharath Hariharan, and Serge Belongie.
\newblock Feature pyramid networks for object detection.
\newblock In \emph{Proceedings of the IEEE conference on computer vision and pattern recognition}, pages 2117--2125, 2017.

\bibitem[Luiten et~al.(2021)Luiten, Osep, Dendorfer, Torr, Geiger, Leal-Taix{\'e}, and Leibe]{HOTA}
Jonathon Luiten, Aljosa Osep, Patrick Dendorfer, Philip Torr, Andreas Geiger, Laura Leal-Taix{\'e}, and Bastian Leibe.
\newblock Hota: A higher order metric for evaluating multi-object tracking.
\newblock \emph{International journal of computer vision}, 129:\penalty0 548--578, 2021.

\bibitem[Millidge et~al.(2021)Millidge, Tschantz, Seth, and Buckley]{NeuralKalman}
Beren Millidge, Alexander Tschantz, Anil Seth, and Christopher Buckley.
\newblock Neural kalman filtering.
\newblock \emph{arXiv preprint arXiv:2102.10021}, 2021.

\bibitem[Murthy et~al.(2020)Murthy, Hashmi, Bokde, and Geem]{murthy2020investigations}
Chinthakindi~Balaram Murthy, Mohammad~Farukh Hashmi, Neeraj~Dhanraj Bokde, and Zong~Woo Geem.
\newblock Investigations of object detection in images/videos using various deep learning techniques and embedded platforms—a comprehensive review.
\newblock \emph{Applied sciences}, 10\penalty0 (9):\penalty0 3280, 2020.

\bibitem[Perlin(1985)]{PerlinNoise}
Ken Perlin.
\newblock An image synthesizer.
\newblock \emph{ACM Siggraph Computer Graphics}, 19\penalty0 (3):\penalty0 287--296, 1985.

\bibitem[Pham et~al.(2020)Pham, Pham, and Dang]{pham2020road}
Vung Pham, Chau Pham, and Tommy Dang.
\newblock Road damage detection and classification with detectron2 and faster r-cnn.
\newblock In \emph{2020 IEEE International Conference on Big Data (Big Data)}, pages 5592--5601. IEEE, 2020.

\bibitem[Ravi et~al.(2024)Ravi, Gabeur, Hu, Hu, Ryali, Ma, Khedr, R{\"a}dle, Rolland, Gustafson, et~al.]{SAM2}
Nikhila Ravi, Valentin Gabeur, Yuan-Ting Hu, Ronghang Hu, Chaitanya Ryali, Tengyu Ma, Haitham Khedr, Roman R{\"a}dle, Chloe Rolland, Laura Gustafson, et~al.
\newblock Sam 2: Segment anything in images and videos.
\newblock \emph{arXiv preprint arXiv:2408.00714}, 2024.

\bibitem[Simonyan et~al.(2013)Simonyan, Vedaldi, and Zisserman]{SaliencyMapsBase}
Karen Simonyan, Andrea Vedaldi, and Andrew Zisserman.
\newblock Deep inside convolutional networks: Visualising image classification models and saliency maps.
\newblock \emph{arXiv preprint arXiv:1312.6034}, 2013.

\bibitem[Szab{\'o} et~al.(2023)Szab{\'o}, Bonaiuti, Ciliberto, and Horv{\'a}th]{szabo2023enhancing}
Gergely Szab{\'o}, Paolo Bonaiuti, Andrea Ciliberto, and Andr{\'a}s Horv{\'a}th.
\newblock Enhancing cell tracking with a time-symmetric deep learning approach.
\newblock \emph{arXiv preprint arXiv:2308.03887}, 2023.

\bibitem[Tarjan(1972)]{DFS}
Robert Tarjan.
\newblock Depth-first search and linear graph algorithms.
\newblock \emph{SIAM journal on computing}, 1\penalty0 (2):\penalty0 146--160, 1972.

\bibitem[Terrell and Scott(1992)]{KDE}
George~R Terrell and David~W Scott.
\newblock Variable kernel density estimation.
\newblock \emph{The Annals of Statistics}, pages 1236--1265, 1992.

\bibitem[Voigtlaender et~al.(2019)Voigtlaender, Krause, Osep, Luiten, Sekar, Geiger, and Leibe]{MOTS}
Paul Voigtlaender, Michael Krause, Aljosa Osep, Jonathon Luiten, Berin Balachandar~Gnana Sekar, Andreas Geiger, and Bastian Leibe.
\newblock Mots: Multi-object tracking and segmentation.
\newblock In \emph{Proceedings of the ieee/cvf conference on computer vision and pattern recognition}, pages 7942--7951, 2019.

\bibitem[Wan and Van Der~Merwe(2000)]{UnscentedKalman}
Eric~A Wan and Rudolph Van Der~Merwe.
\newblock The unscented kalman filter for nonlinear estimation.
\newblock In \emph{Proceedings of the IEEE 2000 adaptive systems for signal processing, communications, and control symposium (Cat. No. 00EX373)}, pages 153--158. Ieee, 2000.

\bibitem[Wu et~al.(2019)Wu, Kirillov, Massa, Lo, and Girshick]{Detectron2}
Yuxin Wu, Alexander Kirillov, Francisco Massa, Wan-Yen Lo, and Ross Girshick.
\newblock Detectron2.
\newblock \url{https://github.com/facebookresearch/detectron2}, 2019.

\end{thebibliography}
}


\end{document}